\documentclass{article}
\usepackage{ijcai19}

\usepackage{times}
\usepackage{soul}
\usepackage{url}
\usepackage[hidelinks]{hyperref}
\usepackage[utf8]{inputenc}
\usepackage[small]{caption}
\usepackage{array}

\usepackage{graphicx}
\usepackage{amsmath}
\usepackage{booktabs}
\usepackage{mathrsfs}
\usepackage{bm,bbm}
\usepackage{amsfonts}
\usepackage{multirow}
\usepackage{bigstrut,bigdelim}
\usepackage{graphicx}  
\usepackage{float}
\usepackage{paralist}
\usepackage{subfigure}
\usepackage{diagbox}

\newcommand{\newcite}[1]{\citeauthor{#1} (\citeyear{#1})}

\title{A Document-grounded Matching Network for Response Selection in Retrieval-based Chatbots}

\author{
Xueliang Zhao\thanks{Equal Contribution.} $^1$
\and
Chongyang Tao\footnotemark[1]$^1$ \and
Wei Wu$^2$ \and
Can Xu$^2$ \and
Dongyan Zhao$^{1,3}$ \And
Rui Yan$^{1,3}$\thanks{Corresponding author: Rui Yan (ruiyan@pku.edu.cn).}~
\affiliations
$^1$Institute of Computer Science and Technology, Peking University, Beijing, China \\
$^2$Microsoft Corporation, Beijing, China\\
$^3$Center for Data Science, Peking University, Beijing, China 
\emails
\{xl.zhao,chongyangtao,zhaody,ruiyan\}@pku.edu.cn, \{wuwei,caxu\}@microsoft.com
}

\begin{document}

\maketitle

\begin{abstract}
We present a document-grounded matching network (DGMN) for response selection that can power a knowledge-aware retrieval-based chatbot system. The challenges of building such a model lie in how to ground conversation contexts with background documents and how to recognize important information in the documents for matching. To overcome the challenges, DGMN fuses information in a document and a context into representations of each other, and dynamically determines if grounding is necessary and importance of different parts of the document and the context through hierarchical interaction with a response at the matching step. Empirical studies on two public data sets indicate that DGMN can significantly improve upon state-of-the-art methods and at the same time enjoys good interpretability.  
\end{abstract}

\section{Introduction}
Human-machine conversation is a long-standing goal of artificial intelligence. Recently, building a chatbot for open domain conversation has gained increasing interest due to both availabilities of a large amount of human conversation data and powerful models learned with neural networks. Existing methods are either retrieval-based or generation-based.  Retrieval-based methods respond to human input by selecting a response from a pre-built index \cite{ji2014information,yan2018coupled}, while generation-based methods synthesize a response with a natural language model \cite{shangL2015neural,li2015diversity}. In this work, we study the problem of response selection for retrieval-based chatbots, since retrieval-based systems are often superior to their generation-based counterparts on response fluency and diversity, are easy to evaluate.

\begin{table}[]
\resizebox{0.48\textwidth}{!}{
\begin{tabular}{|c|l|}
\hline
\multirow{4}{*}{\textbf{A's profile}} & {trying new recipes makes me happy.} \\ 
& {i feel like i need to exercise more.}   \\ 
& {i am an early bird , while my significant other is a night owl.} \\ 
& {i am a kitty owner.}  \\ \hline
\multirow{4}{*}{\textbf{B's profile}} & i might actually be a mermaid.  \\
& i use all of my time for my education.                               \\
& i am very sociable and love those close to me.                       \\
& i enjoy swimming in the ocean , i feel in tune with its inhabitants. \\ \hline
\multirow{3}{*}{\textbf{Context}} & {\textbf{A}: hi how are you today} \\
& {\textbf{B}: i am good . how are you ?}             \\
& {\textbf{A}: pretty good where do you work ?}        \\ \hline
\textbf{True response} & {i do not work , i am a full time student . what about you?} \\
\textbf{False response} & {i have been working as a salesman for more than $10$ years.}   \\ \hline
\end{tabular}
}
\caption{An example of document-grounded dialogue}
\label{tab:example}
\end{table}

A key step in response selection is measuring the matching degree between a context (a message with a few turns of conversation history) and a response candidate.  Existing methods \cite{wu2017sequential,zhou2018multi} have achieved impressive performance on benchmarks \cite{lowe2015ubuntu,wu2017sequential}, but responses are selected solely based on conversation history.   Human conversations, on the other hand, are often grounded in external knowledge. For example, in Reddit, discussion among users is usually along the document posted at the beginning of a thread which provides topics and basic facts for the following conversation.  Lack of knowledge grounding has become one of the major gaps between the current open domain dialog systems and real human conversations.  As a step toward bridging the gap, we investigate knowledge-grounded response selection in this work and specify the knowledge as unstructured documents that are common sources in practice.  The task is that given a document and a conversation context based on the document, one selects a response from a candidate pool that is consistent and relevant with both the conversation context and the background document. Table \ref{tab:example} shows an example from PERSONA-CHAT, a data set released recently by Facebook \cite{zhang2018personalizing}, to illustrate the task: given two speakers' profiles as documents and a conversation context, one is required to distinguish the true response from the false ones\footnote{For space limitation, we only show one false response here.}.

Intuitively, both documents and conversation contexts should participate in matching.   Since documents and contexts are highly asymmetric in terms of information they convey, and there exists complicated dependency among sentences in the documents and utterances in the contexts, challenges of the task include (1) how to ground conversation contexts with documents given that utterances in the contexts are not always related to the documents due to the casual nature of open domain conversation (e.g., the greetings in Table \ref{tab:example}); (2) how to comprehend documents with conversation contexts when information in the documents are rather redundant for proper response recognition (e.g., the description regarding to B's hobby in her profile in Table \ref{tab:example}); and (3) how to effectively leverage both information sources to perform matching.  To overcome the challenges, we propose a document-grounded matching network (DGMN). DGMN encodes sentences in a document, utterances in a conversation context, and a response candidate through self-attention, and models context grounding and document comprehension by constructing a document-aware context representation and a context-aware document representation via an attention mechanism. With the rich representations, DGMN distills matching information from each utterance-response pair and each sentence-response pair, where whether an utterance needs grounding, which parts of the document are crucial for grounding and matching, and which parts of the context are useful for representing the document are dynamically determined by a hierarchical interaction mechanism.  The final matching score is defined as an aggregation of matching signals from all pairs. 

We conduct experiments on two public data sets: the PERSONA-CHAT data \cite{zhang2018personalizing} and the CMU Document Grounded Conversation (CMUDoG) data \cite{zhou2018dataset}. Evaluation results indicate that on both data sets, DGMN can significantly outperform state-of-the-art methods. Compared with Transformer, the best performing baseline on both data, absolute improvements from DGMN on $r@1$ (hits@1) are more than $13$\% on the PERSONA-CHAT data and more than $5$\% on the CMUDoG data. Through both quantitative and qualitative analysis, we also demonstrate the effect of different representations to matching and how DGMN grounds conversation contexts with documents.  

Our contributions in this work are three-fold: (1) proposal of a document-grounded matching network that performs response selection according to both conversation contexts and background knowledge; (2) empirical verification of the effectiveness of the proposed model on two public data sets; and (3) new state-of-the-art on the PERSONA-CHAT data without any pre-training on external resources.

\section{Document-Grounded Matching Network}
In this section, we first formalize the document-grounded matching problem, and then introduce our model from an overview to details of components. 

\subsection{Problem Formalization}
Suppose that we have a data set $\mathcal {D} = \{(D_i, c_i, y_i, r_i)\}_{i=1}^N$ where $D_i=\{d_{i, 1}, \cdots, d_{i, m_i}\}$ is a document that serves as background knowledge for conversation with $d_{i,k}$ the $k$-th sentence, $c_i=\{u_{i, 1}, \cdots, u_{i, n_i}\}$ is a conversation context following $D_i$ with $u_{i,k}$ the $k$-th utterance, $r_i$ is a response candidate, and $y_i \in \{0, 1\}$ is a label with $y_i=1$ indicating that $r_i$ is a proper response given $c_i$ and $D_i$, otherwise $y_i=0$. The task is to learn a matching model $g(\cdot,\cdot,\cdot)$ from $\mathcal{D}$, and thus for a new triple $(D,c,r)$, $g(D,c,r)$ returns the matching degree between $c$ and $r$ under $D$.

\subsection{Model Overview}
\begin{figure*}
    \centering
    \includegraphics[width=0.74\linewidth]{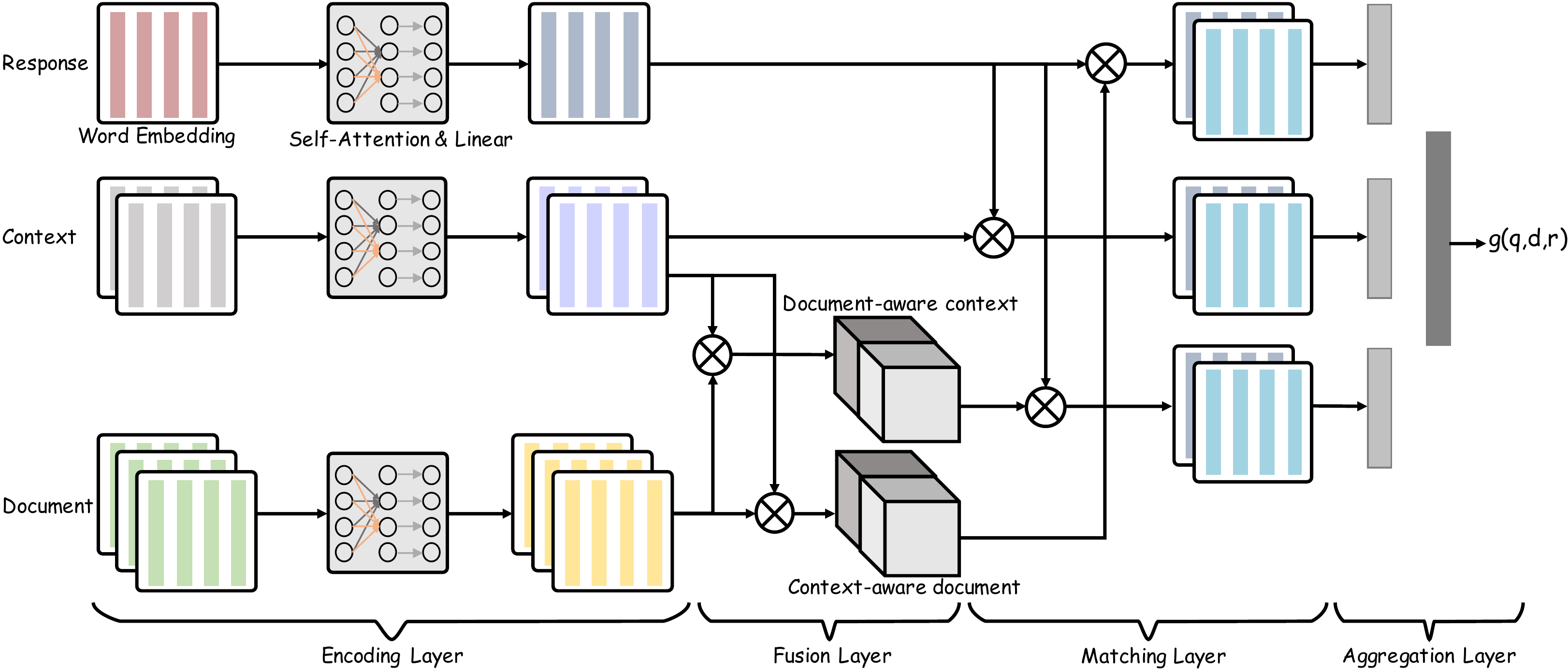}
    \caption{Architecture of the document-grounded matching network.}
    \label{fig:model}
\end{figure*}

We define $g(D,c,r)$ as a document-grounded matching network. Figure \ref{fig:model} illustrates the architecture of the model. In brief, DGMN consists of an encoding layer, a fusion layer, a matching layer, and an aggregation layer. The encoding layer represents $D$, $c$, and $r$ via self-attention, and feeds the representations to the fusion layer where $D$ and $c$ are fused into the representations of each other as a document-aware context representation and a context-aware document representation.  Based on the representations given by the first two layers, the matching layer then lets each utterance in $c$ and each sentence in $D$ interact with $r$, and distills matching signals from the interaction. Matching signals in all pairs are finally aggregated as a matching score in the aggregation layer.

\subsection{Model Details}
We elaborate each layer of the document-grounded matching network in this section. 

\subsubsection{Encoding Layer}
Given an utterance $u_i$ in a context $c$, a sentence $d_j$ in a document $D$, and a response candidate $r$, the model first embeds $u_i$, $d_i$, and $r$ as $\mathbf{E}_{u_i} = [\mathbf{e}_{{u_i},1}, \cdots, \mathbf{e}_{{u_i}, l_u}]$, $\mathbf{E}_{d_j} = [\mathbf{e}_{{d_j},1}, \cdots, \mathbf{e}_{{d_j}, l_d}]$ and $\mathbf{E}_r = [\mathbf{e}_{r,1}, \cdots ,\mathbf{e}_{r,l_r}]$ respectively by looking up a shared embedding table pre-trained with Glove~\cite{pennington2014glove} on the training data $\mathcal{D}$, where $\mathbf{e}_{{u_i},k}$, $\mathbf{e}_{{d_j},k}$ and  $\mathbf{e}_{r,k}$ are representations of the $k$-th words in $u_i$, $d_j$ and $r$ respectively, and $l_u$, $l_r$, and $l_d$ are lengths of the three sequences. $\mathbf{E}_{u_i}$, $\mathbf{E}_{d_j}$ and $\mathbf{E}_r$ are then processed by an attentive module to encode long-term dependency among words into the representations. 

The attentive module simplifies the multi-head attention module in Transformer \cite{vaswani2017attention}, and consists of a scaled dot-product attention component and a feed-forward component. Without loss of generality, let $\mathbf{Q} \in \mathbb{R}^{n_Q \times d}$, $\mathbf{K} \in \mathbb{R}^{n_K \times d}$, and $\mathbf{V} \in \mathbb{R}^{n_V \times d}$ denote embedding matrices of a query, a key, and a value respectively, where $n_Q$, $n_K$, and $n_V$ are numbers of words in the input sequences, and $d$ stands for embedding size. The scaled dot-product attention component is then defined as:
\begin{equation}
\begin{aligned} \label{eq:dot}
    \text{Attention}(\mathbf{Q},\mathbf{K},\mathbf{V}) = \text{softmax}(\mathbf{Q} \mathbf{K}^{T}/\sqrt{d})\mathbf{V}.
\end{aligned}
\end{equation}

Intuitively, each entry of $\mathbf{V}$ is weighted by a relevance score defined by the similarity of an entry of $\mathbf{Q}$ and an entry of $\mathbf{K}$, and then an updated representation of $\mathbf{Q}$ is formed by linearly combining the entries of $\mathbf{V}$ with the weights. In practice, we often let $\mathbf{K}=\mathbf{V}$, and thus $\mathbf{Q}$ is represented by similar entries of $\mathbf{V}$. The feed-forward component takes $\text{Attention}(\mathbf{Q},\mathbf{K},\mathbf{V})$ as input, and transforms it to a new representation by two non-linear projections. A residual connection~\cite{he2016deep} and a row-wise normalization~\cite{ba2016layer} are applied to the result of each projection. For ease of presentation, we denote the whole attentive module as $f_{\text{ATT}}(\mathbf{Q}, \mathbf{K}, \mathbf{V})$. $u_i$, $d_j$ and $r$ are then represented by attending to themselves through $f_{\text{ATT}}(\cdot, \cdot, \cdot)$:
\begin{align}
    \mathbf{U}_i &= f_{\text{ATT}}(\mathbf{E}_{u_i}, \mathbf{E}_{u_i}, \mathbf{E}_{u_i}) \\
    \mathbf{D}_j &= f_{\text{ATT}}(\mathbf{E}_{d_j}, \mathbf{E}_{d_j}, \mathbf{E}_{d_j}) \\
    \mathbf{R} &= f_{\text{ATT}}(\mathbf{E}_r, \mathbf{E}_r, \mathbf{E}_r).
\end{align}

\subsubsection{Fusion Layer}
The fusion layer grounds the conversation context by the document and fuses the information of the context into the document, which results in a document-aware context representation and a context-aware document representation. Formally, the document-aware representation of $u_i$ is given by  $\mathbf{\hat{U}}_{i}=[\mathbf{\hat{U}}_{i,1}, \cdots, \mathbf{\hat{U}}_{i,m}]$, where $m$ is the number of sentences in the document, and $\forall j \in \{1,\ldots, m\}$, $\mathbf{\hat{U}}_{i,j}$ can be formulated as 
\begin{align}
    \mathbf{\hat{U}}_{i,j} &= f_{\text{ATT}}(\mathbf{U}_i, \mathbf{D}_j, \mathbf{D}_j).
\end{align}

Similarly, the context-aware representation of $d_j$ is defined as $\mathbf{\hat{D}}_{j}=[\mathbf{\hat{D}}_{j,1}, \cdots, \mathbf{\hat{D}}_{j,n}]$, where $n$ is the number of utterances in the context, and $\forall i \in \{1,\ldots, n\}$, $\mathbf{\hat{D}}_{j,i}$ is calculated by
\begin{align}
    \mathbf{\hat{D}}_{j,i} &= f_{\text{ATT}}(\mathbf{D}_j, \mathbf{U}_i, \mathbf{U}_i).
\end{align}

In $\mathbf{\hat{U}}_{i,j}$, information in $d_j$ provides grounding to $u_i$, and correlations between $d_j$ and $u_i$ will be distilled to enhance the original representation of $u_i$. The grounding is performed on a sentence-level rather than on a document-level (i.e., attention with a document vector). This is motivated by the intuition that sentences in a document are differentially important to represent the semantics of an utterance in a context, and the importance should be dynamically recognized through interaction with a response in the matching step.  In a similar sense, by letting $d_j$ attend to $u_i$ in $\mathbf{\hat{D}}_{j,i}$
we attempt to highlight important parts of $d_j$ through their correlation with $u_i$, and thus achieve better document understanding in matching.

As we have analyzed before, utterances in a context are not always related to the background document in chat. To model this intuition, we append $\mathbf{U}_i$ to $\mathbf{\hat{U}}_{i}$ as $\mathbf{\tilde{U}}_{i}=[\mathbf{U}_i, \mathbf{\hat{U}}_{i,1}, \cdots, \mathbf{\hat{U}}_{i,m}]$ and determine if an utterance needs grounding with the guide of response $r$ in the following matching layer. Ideally, if an utterance does not need grounding, then only $\mathbf{U}_i$ should participate in matching since other entries of $\mathbf{\tilde{U}}_{i}$ are noisy. The weights of the entries of $\mathbf{\tilde{U}}_{i}$ will be learned from training data.  

\subsubsection{Matching Layer}
The matching layer pairs $\mathbf{U}_i$, $\mathbf{\tilde{U}}_{i}$, $\mathbf{\hat{D}}_{j}$ with $\mathbf{R}$ as $\{\mathbf{U}_i,\mathbf{R}\}$, $\{\mathbf{\tilde{U}}_{i},\mathbf{R}\}$ and $\{\mathbf{\hat{D}}_{j},\mathbf{R}\}$ respectively, and extracts matching information from the pairs. Different from existing matching models that are solely based on conversation contexts, $\mathbf{\tilde{U}}_{i}$ and $\mathbf{\hat{D}}_{j}$ now contain grounding information from multiple sentences (utterances). Thus, the model needs to dynamically select important sentences (utterances) for grounding and even determine if grounding is necessary. To tackle the new challenges, we propose a hierarchical interaction mechanism. Take $\{\mathbf{\tilde{U}}_{i}, \mathbf{R}\}$ as an example. For ease of presentation, we define $\mathbf{U}_i=\mathbf{\hat{U}}_{i,0}$. Let $\mathbf{r}_j$ denote the $j$-th entry of $\mathbf{R}$, then the first level interaction of $\mathbf{\tilde{U}}_{i}$ and $\mathbf{R}$ happens between $\mathbf{r}_j$ and each $\mathbf{\hat{U}}_{i,k}$, $\forall k\in \{0,\ldots, m\}$, and transforms $\mathbf{\hat{U}}_{i,k}$ into $h_{i,j,k}$ through 
\begin{align} 
\label{eq:atten1}
    \omega_{i, j, k, t} &= \mathbf{v}_a^{\top} \text{tanh}(\mathbf{w}_{a} [\mathbf{\hat{u}}_{i,k,t}; \mathbf{r}_{j}] + \mathbf{b}_{a}),\\
    \label{eq:atten2}
    \alpha_{i, j, k, t} &= \frac{\exp(\omega_{i,j,k,t})}{\sum_{t=1}^{l_u} \exp(\omega_{i,j,k,t})},\\
    \label{eq:atten3} 
    h_{i, j, k} &= \sum\nolimits_{t=1}^{l_u} \alpha_{i,j,k,t} {\mathbf{\hat{u}}_{i,k,t}},
\end{align}
where $\mathbf{\hat{u}}_{i,k,t}$ is the $t$-th entry of $\mathbf{\hat{U}}_{i,k}$, and $\mathbf{w}_{a}$, $\mathbf{v}_{a}$, and $\mathbf{b}_{a}$ are parameters. Through Eq. (\ref{eq:atten3}), the first level interaction tries to play emphasis on important words in each $\mathbf{\hat{U}}_{i,k}$ with respect to $\mathbf{r}_j$. The second level interaction of $\mathbf{\tilde{U}}_{i}$ and $\mathbf{R}$ then summarizes $[h_{i,j,0}, \ldots, h_{i,j,m}]$ as $h_{i,j}$ by
\begin{align} 
    \label{eq:second_atten1}
    \omega'_{i,j,k} &= \mathbf{v'}_a^{\top} \text{tanh}(\mathbf{w'}_{a} [h_{i,j,k}; \mathbf{r}_{j}] + \mathbf{b'}_{a}),\\
    \label{eq:second_atten2}
    \alpha'_{i,j,k} &= \frac{\exp(\omega'_{i,j,k})}{\sum_{k=0}^{m} \exp(\omega'_{i,j,k})},\\
    \label{eq:second_atten3}
    h_{i, j} &= \sum\nolimits_{k=0}^{m} \alpha'_{i,j,k} {h_{i,j,k}},
\end{align}
where $\mathbf{w'}_{a}$, $\mathbf{v'}_{a}$, and $\mathbf{b'}_{a}$ are parameters. In the second level interaction, sentences in the document that can bring valuable grounding information for matching will play an important role in the formation of $h_{i,j}$. As a special case, when $\alpha'_{i,j,0}$ is much bigger than other weights, the model judges that $u_i$ does not need grounding from the document. Finally, matching information between $\mathbf{\tilde{U}}_{i}$ and $\mathbf{R}$ is stored in a matrix $\mathbf{\tilde{M}}_i = [\mathbf{m}_{i,1} , \cdots, \mathbf{m}_{i,l_r}]$. $\forall j \in \{1,\ldots, l_r\}$,  $\mathbf{m}_{i,j}$ is calculated by
\begin{equation}  
\label{eq:interfunction}
\mathbf{m}_{i,j}=\text{ReLU}(\mathbf{w}_p \begin{bmatrix} (h_{i,j} - \mathbf{r}_j) \odot (h_{i,j} - \mathbf{r}_j) \\ h_{i,j} \odot \mathbf{r}_j \end{bmatrix}  + \mathbf{b}_p),
\end{equation}
where $\mathbf{w}_{p}$ and $\mathbf{b}_{p}$ are  parameters, and $\odot$ refers to element-wise multiplication. 

Following the same procedure, we obtain $\mathbf{\hat{M}}_j$ as a matching matrix for $\{\mathbf{\hat{D}}_{j},\mathbf{R}\}$ where utterances in the context that are helpful for representing $d_j$ are highlighted by $r$. Since $\mathbf{U}_i$ is only made up of word representations (i.e., one-layer structure), the matching matrix  $\mathbf{M}_i$ for $\{\mathbf{U}_i,\mathbf{R}\}$ is calculated by one level interaction parameterized in a similar way as Eq. (\ref{eq:atten1})-(\ref{eq:atten3}) and the same function as Eq. (\ref{eq:interfunction}).

\subsubsection{Aggregation Layer and Learning Method}
The aggregation layer accumulates matching signals in $\{\mathbf{M}_i\}_{i=1}^n$, $\{\mathbf{\tilde{M}}_i\}_{i=1}^n$, and $\{\mathbf{\hat{M}}_j\}_{j=1}^m$ as a matching score for $(D,c,r)$. Specifically, we construct a tensor from $\{\mathbf{M}_i\}_{i=1}^n$, and then apply a convolutional neural network \cite{ji20103d} to the tensor to calculate a matching vector $\mathbf{t}$. Similarly, we have 
matching vectors $\mathbf{\hat{t}}$ and $\mathbf{\tilde{t}}$ for $\{\mathbf{\hat{M}}_j\}_{j=1}^m$ and $\{\mathbf{\tilde{M}}_i\}_{i=1}^n$, respectively. The matching function $g(D, c, r)$ is defined as
\begin{equation} \label{matchingmodel}
    g(D, c, r) = \sigma([\mathbf{t}; \mathbf{\hat{t}}; \mathbf{\tilde{t}}] \mathbf{w}_o + \mathbf{b}_o),
\end{equation}
where $\mathbf{w}_o$ and $\mathbf{b}_o$ are wights, and $\sigma(\cdot)$ is a sigmoid function.

Parameters of $g(D, c, r)$ are estimated from the training data $\mathcal{D}$ by minimizing the following objective:
\begin{equation} \small
- \sum_{i=1}^N \Big( y_i \log g(D_i, c_i, r_i)  +  (1-y_i)  \log (1- g(D_i, c_i, r_i)) \Big).
\end{equation}

\section{Experiments}
We test our model on two public data sets. 
\subsection{Experimental Setup}
The first data we use is the PERSONA-CHAT data set published in~\cite{zhang2018personalizing}. The data is collected by requiring two workers on Amazon Mechanical Turk to chat with each other according to their assigned profiles. Each profile is presented in a form of a document with an average of $4.49$ sentences. The profiles define speakers' personas and provide characteristic knowledge for dialogues. For each dialogue, there are both original profiles and revised profiles that are rephrased from the original ones by other crowd workers to force models to learn more than simple word overlap. A revised profile shares the same number of sentences with its original one, and on average, there are $7.33$ words per sentence in the original profiles and $7.32$ words per sentence in the revised ones. The data is split as a training set, a validation set, and a test set by the publishers. In all the three sets, $7$ turns before an utterance are used as conversation history, and the next turn of the utterance is treated as a positive response candidate. Besides, each utterance is associated with $19$ negative response candidates that are randomly sampled by the publishers. More statistics of the three sets are shown in Table \ref{tbl:stat}. Following the insights in \cite{zhang2018personalizing}, we train models using revised profiles and test the models with both original and revised profiles. 

In addition to PERSONA-CHAT, we also conduct experiments with CMUDoG data set published recently in~\cite{zhou2018dataset}. Conversations in the data are collected from workers on Amazon Mechanical Turk and are based on movie-related wiki articles in two scenarios. In the first scenario, only one worker has access to the provided document, and he/she is responsible for introducing the movie to the other worker; while in the second scenario, both workers know the document and they are asked to discuss the content of the document. Since the data size for an individual scenario is small, we merge the data of the two scenarios in the experiments and filter out conversations less than $4$ turns to avoid noise.  Each document consists of $4$ sections and these sections are shown to the workers one by one every $3$ turn (the first section lasts $6$ turns due to initial greetings). On average, each section contains $8.22$ sentences and $27.86$ words per sentence. The data has been divided into a training set, a validation set, and a test set by the publishers. In each set, we take $2$ turns before an utterance as conversation history and the next turn of the utterance as a positive response candidate. Since the data does not contain negative examples, we randomly sample $19$ negative response candidates for each utterance from the same set. Detailed statistics of the data is given in Table \ref{tbl:stat}. 

We employ $r@k$ as evaluation metrics where $k\in \{1,2,5\}$. For a single context, if the only positive candidate is ranked within top $k$ positions, then $r@k=1$, otherwise, $r@k=0$. The final value of the metric is an average over all contexts in test data. Note that in PERSONA-CHAT, $r@1$ is equivalent to hits@1 which is the metric used by \cite{zhang2018personalizing} for model comparison. 

\begin{table}[t!]
\centering
\resizebox{0.48\textwidth}{!}{
\begin{tabular}{|l|c|c|c|c|c|c|}
\hline
\multirow{2}{*}{Statistics}
& \multicolumn{3}{c|}{PERSONA-CHAT} & \multicolumn{3}{c|}{CMUDoG} \\ \cline{2-7}
& Train  & Val   & Test & Train  & Val   & Test  \\ \hline
\# of conversations       & 8939   & 1000  & 968  & 2881  & 196   & 537  \\ \hline
\# of turns               & 65719  & 7801  & 7512 & 36159 & 2425  & 6637  \\ \hline
Av\_turns / conversation   & 7.35   & 7.80  & 7.76 & 12.55 & 12.37  & 12.36  \\ \hline
Av\_length of utterance     & 11.67 & 11.94 & 11.79 & 18.64 & 20.06 & 18.11  \\ \hline
\end{tabular}
}
\caption{Statistics of the two data sets.}
\label{tbl:stat}
\end{table}

\begin{table*}[t!]
\centering
\resizebox{0.73\textwidth}{!}{
\begin{tabular}{|l|c|c|c|c|c|c|c|c|c|}
    \hline
     \multirow{3}{*}{\backslashbox{Models}{Metrics}}  & \multicolumn{6}{c|}{PERSONA-CHAT} & \multicolumn{3}{c|}{\multirow{2}{*}{CMUDoG}} \\ \cline{2-7}
     &  \multicolumn{3}{c|}{Original Persona}   & \multicolumn{3}{c|}{Revised Persona}  & \multicolumn{3}{c|}{} \\ \cline{2-10}
                    & $r@1$ & $r@2$ & $r@5$ & $r@1$ & $r@2$ & $r@5$ & $r@1$ & $r@2$ & $r@5$ \\ \hline
    Starspace~\cite{wu2018starspace}       & 49.1  & 60.2     & 76.5     & 32.2     & 48.3     & 66.7   & 50.7     & 64.5     & 80.3    \\ 
    Profile Memory~\cite{zhang2018personalizing}  & 50.9  & 60.7     & 75.7     & 35.4     & 48.3     & 67.5  & 51.6  & 65.8    & 81.4     \\
    KV Profile Memory~\cite{zhang2018personalizing}  & 51.1 & 61.8    & 77.4     & 35.1     & 45.7     & 66.3  & 56.1     & 69.9     & 82.4      \\ 
    Transformer~\cite{mazare2018training} & 54.2 & 68.3    & 83.8     & 42.1     & 56.5     & 75.0  & 60.3    & 74.4     & 87.4 \\ \hline
    DGMN             & \textbf{67.6} & \textbf{80.2} & \textbf{92.9}  & \textbf{58.8} & \textbf{62.5} & \textbf{87.7}  & \textbf{65.6} & \textbf{78.3} & \textbf{91.2} \\ \hline
    DGMN($\mathbf{t}$)  & 51.8  & 66.1  & 83.3  & 51.8  & 66.1 & 83.3  & 55.6 & 69.4 & 85.4  \\ 
    DGMN($\mathbf{t}$+$\mathbf{\tilde{t}}$) &  66.3 & 78.9 & 91.7 & 57.0 & 71.2 & 86.9 & 64.5 & 78.2 & 90.8 \\
    DGMN($\mathbf{t}$+$\mathbf{\tilde{t}}$-NoGround)   & 64.2  & 77.8  & 91.3 & 55.8 & 70.1 & 86.2 & 63.5 & 76.8 & 90.8  \\ \hline
    \end{tabular}
}
\caption{Evaluation results on the test sets of the PERSONA-CHAT data and the CMUDoG data. Numbers in bold mean that improvement over the best baseline is statistically significant (t-test, $p$-value $<0.01$).}
\label{tab:main_results}
\end{table*}

\subsection{Baseline Models}
The following models are selected as baselines. These models are the ranking models in \cite{zhang2018personalizing} and \cite{mazare2018training} which perform much better than the generative models in \cite{zhang2018personalizing} on the PERSONA-CHAT data.

\textbf{Starspace}: a supervised model in~\cite{wu2018starspace} that learns the similarity between a conversation context and a response candidate by optimizing task-specific embedding via the margin ranking loss. The similarity is measured by the cosine of the sum of word embeddings. Documents are concatenated to conversation contexts. 
 
\textbf{Profile Memory}: the model in~\cite{zhang2018personalizing} that lets a conversation context attend over the associated document to produce a vector which is then combined with the context. Cosine is used to measure the similarity between the output context representation and a response candidate. 

\textbf{KV Profile Memory}: the best performing model in~\cite{zhang2018personalizing} which considers keys as dialogue history and values as the next dialogue utterances and uses a conversation context as input to perform attention over the keys in addition to the documents. The past dialogues are stored in memory to help influence the prediction for the current conversation.

\textbf{Transformer}: a variant of the model proposed by \newcite{vaswani2017attention} for machine translation. The model exhibits state-of-the-art performance on the 
PERSONA-CHAT data as reported in \cite{mazare2018training}. 

All baseline models are implemented with the code shared at \url{https://github.com/facebookresearch/ParlAI/tree/master/projects/personachat} and tuned on the validation sets. We make sure that the baselines achieve the performance on the PERSONA-CHAT data as reported in \cite{zhang2018personalizing} and \cite{mazare2018training}. Note that we do not include models pre-trained from large-scale external resources, such as the FT-PC model in \cite{mazare2018training}, as baselines, since the comparison is unfair. On the other hand, it is interesting to study if pre-train the proposed model on those large-scale external data (e.g., the Reddit data in \cite{mazare2018training} with over 5 million personas spanning more than 700 million conversations) can further improve its performance. We leave the study as future work.   

\subsection{Implementation Details}
We set the size of word embedding as $300$. In PERSONA-CHAT, the number of sentences per document is limited to $5$ (i.e., $m \leq 5$). For each sentence in a document, each utterance in a context, and each response candidate, if the number of words is less than $20$, we pad zeros, otherwise, we keep the latest $20$ words (i.e., $l_u=l_r=l_d=20$). In CMUDoG, we set $m \leq 20$ and $l_u=l_r=l_d=40$ following the same procedure.  In the matching layer of DGMN, the number of filters of CNN is set as $16$, and the window sizes of convolution and pooling are both $3$. All models are learned using Adam~\cite{kingma2014adam} optimizer with a learning rate of $0.0001$. 
In training, we choose $32$ as the size of mini-batches.
Early stopping on validation data is adopted as a regularization strategy.

\subsection{Evaluation Results}
Table~\ref{tab:main_results} reports evaluation results on the two data sets. We can see that on both data sets, DGMN outperforms all baselines over all metrics, and the improvement is statistically significant (t-test, $p$-value $<0.01$). Improvement from DGMN over Transformer on the CMUDoG data is smaller than that on the PERSONA-CHAT data. The reason might be that Transformer can benefit from the wiki documents in CMUDoG that are longer and contain richer semantics than those handcrafted ones in PERSONA-CHAT.

\begin{figure*}[h!]
  \centering
  \subfigure[\scriptsize{$u$-$d_1$}] { 
    \includegraphics[height=2.5cm]{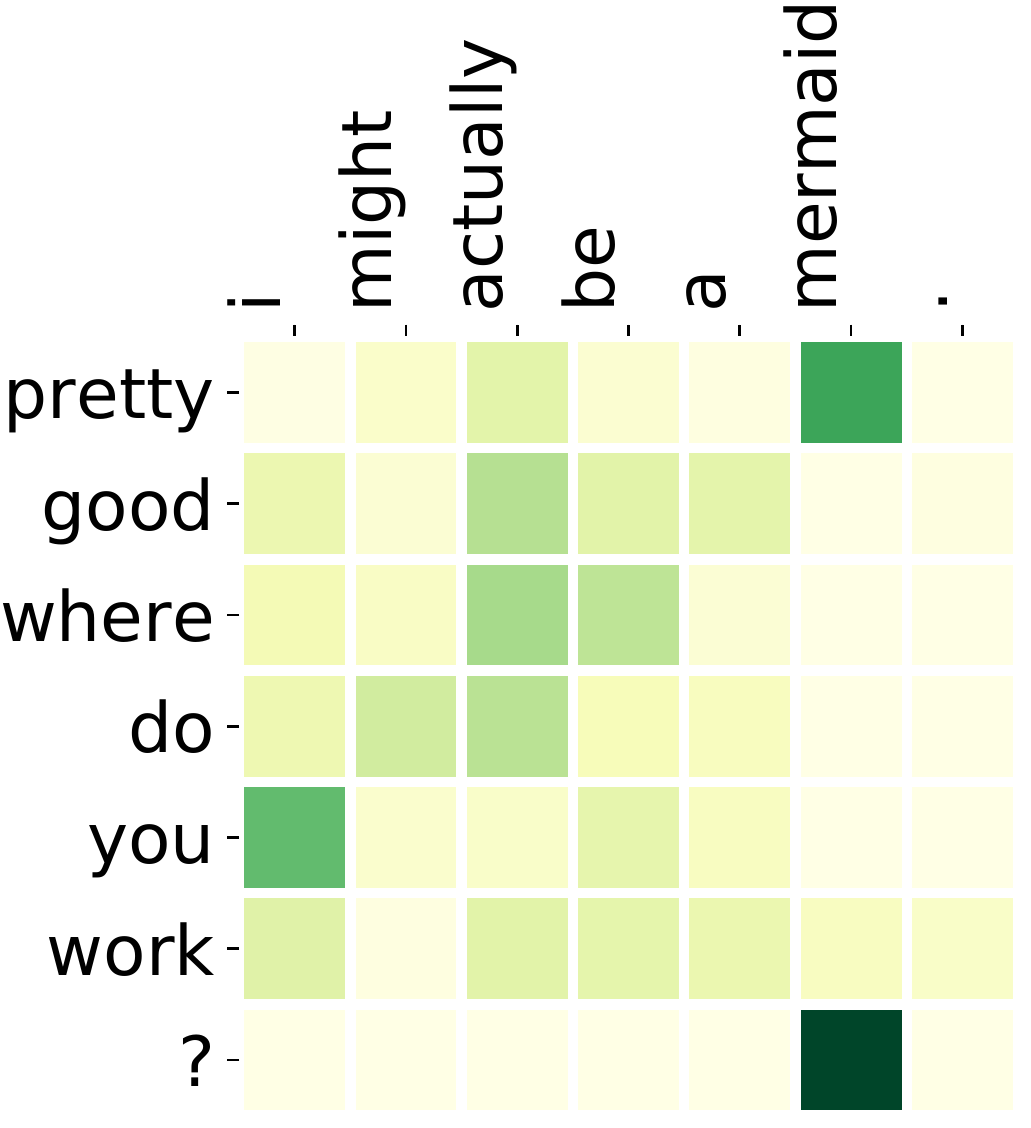}
  } \hspace{-2mm}
  \subfigure[\scriptsize{$u$-$d_2$}] { 
    \includegraphics[height=2.6cm]{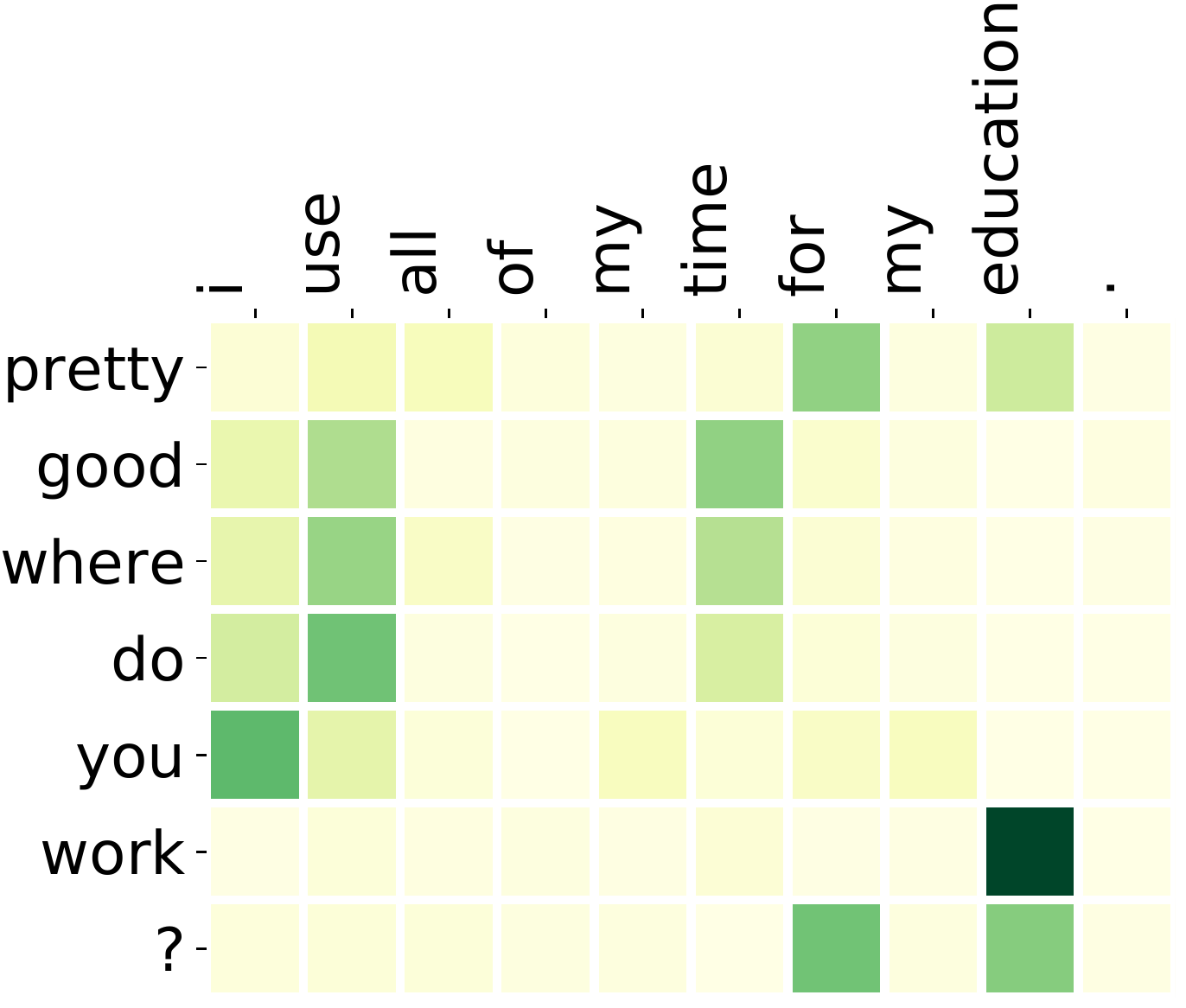}
  } \hspace{-2mm}
  \subfigure[\scriptsize{$u$-$d_3$}] {
    \includegraphics[height=2.5cm]{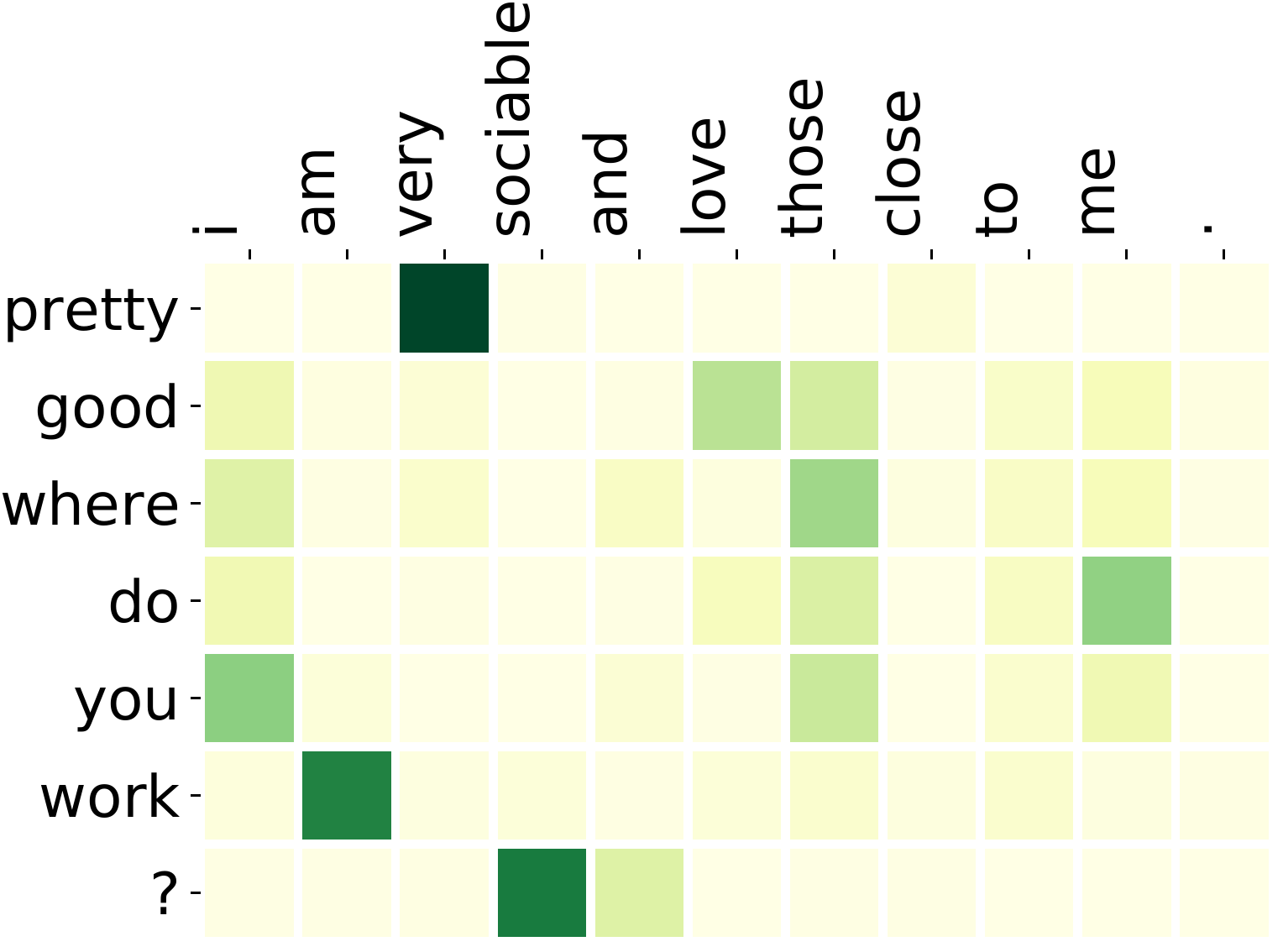}
  } \hspace{-2mm}
  \subfigure[\scriptsize{$u$-$d_4$}] { 
    \includegraphics[height=2.7cm]{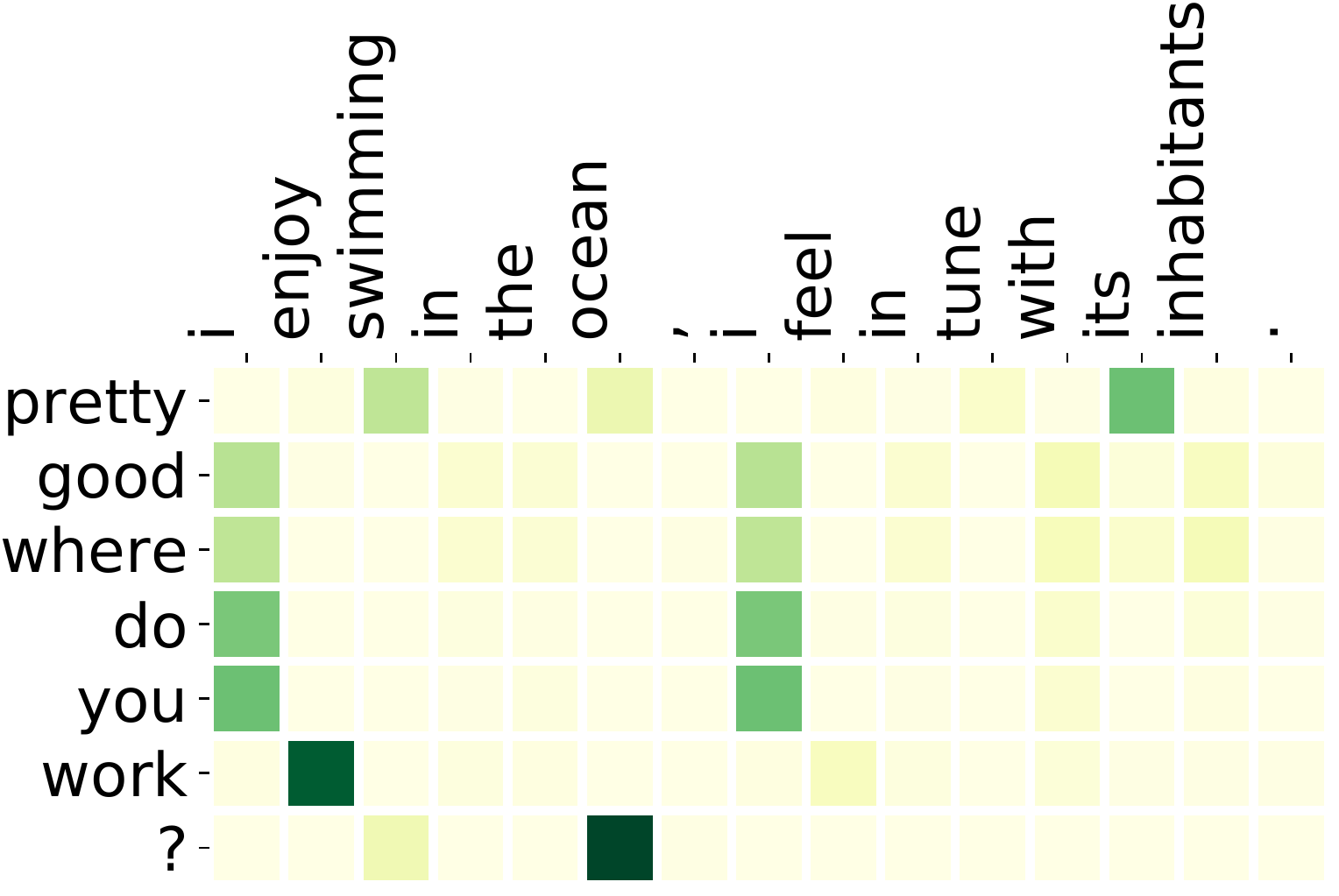}
  } \hspace{-2mm}
  \subfigure[{\scriptsize{Interaction weight}}] { 
    \includegraphics[height=2cm]{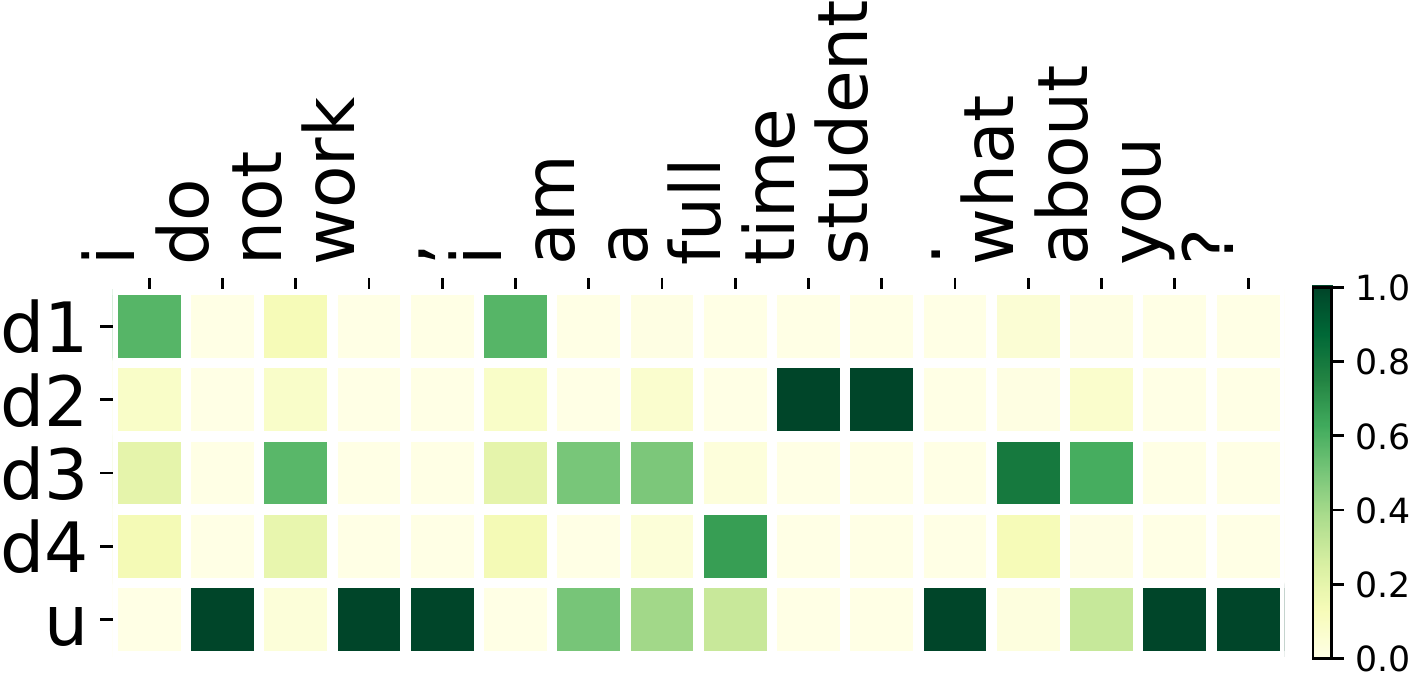}
  }
  \caption{Visualization of context grounding. The first four graphs illustrate attention between the last utterance of the context and each sentence in the document. The last one shows $\alpha'_{i,j,k}$ in interaction.}
  \label{fig:visulization}
\end{figure*}

\begin{table*}[t!]
\resizebox{\textwidth}{!}{
\begin{tabular}{|l|c|c|c|c|c|c|c|c|c|c|c|c|}
\hline
 & \multicolumn{4}{c|}{Original Persona} & \multicolumn{4}{c|}{Revised Persona} & \multicolumn{4}{c|}{CMUDoG}     \\ \cline{1-13}
Doc Length        & (0,25{]} & (25,30{]} & (30,35{]} & (35,52{]} & (0,30{]} & (30, 35{]} & (35, 40{]} & (40,55{]} & (0, 150{]} & (150, 250{]} & (250, 350{]} & (350,515{]} \\ \hline
Case Number       & 1019    & 2099      & 2197      & 2197  & 2419    & 2161       & 1558       & 1374  & 1921      & 2528         & 980          & 1208   \\ \hline
r@1               & 67.1    & 68.8      & 67.5      & 66.7  & 57.8    & 59.7       & 59.1       & 59.0  & 64.0      & 65.8         & 66.2         & 67.4   \\ \hline
\end{tabular}
}
\caption{Performance of DGMN across different length of grounded documents on all data sets.}
\label{tab:doc_lens}
\end{table*}

\subsection{Discussions}
In this section, we investigate how different representations affect the performance of DGMN by an ablation study, visualize the example in Table \ref{tab:example} to illustrate how contexts are grounded by documents in DGMN, and check how the performance of DGMN changes with respect to document length.  

\paragraph{Ablation Study.}  
First, we calculate a matching score only with the self-attention based context representation and response representation and denote the model as DGMN($\mathbf{t}$) which means only $\mathbf{t}$ is kept in Eq. (\ref{matchingmodel}). Then, we take the document-aware context representation into account, and denote the model as DGMN($\mathbf{t}$+$\mathbf{\tilde{t}}$) in which both  $\mathbf{t}$ and $\mathbf{\tilde{t}}$ are used in Eq. (\ref{matchingmodel}). Based on $\mathbf{t}$+$\mathbf{\tilde{t}}$, we further examine if the special configuration for utterances that do not need grounding matters to the performance of DGMN by removing $\mathbf{U}_i$ from $\mathbf{\tilde{U}}_{i}$. The model is denoted as DGMN($\mathbf{t}$+$\mathbf{\tilde{t}}$-NoGround). Finally, the context-aware document representation is considered, and we have the full model of DGMN. Table \ref{tab:main_results} reports evaluation results on the two data sets. We can conclude that (1) all representations are useful for matching; (2) some effect of the context-aware document representation might be covered by the document-aware context representation, as adding the former after the latter does not bring much gain; and (3) although simple, the special configuration for utterances that do not need grounding cannot be removed from DGMN.

\paragraph{Visualization.}
Second, to further understand how DGMN performs context grounding, we visualize the attention weights in formation of the document-aware context representation (i.e., $\mathbf{\hat{U}}_{i,j}$) and the weights in the second level of interaction (i.e., $\alpha'_{i,j,k}$ in Eq. (\ref{eq:second_atten2})) with the example in Table \ref{tab:example} in Introduction. Due to space limitation, we only visualize the last utterance of the context. Figure~\ref{fig:visulization} shows the results. It is interesting to see that words like ``work'' and ``education'' are highly correlated in the graph, and at the same time, weights between the utterance and irrelevant sentences in the profile, such as ``I am very social and love those close to me'', are generally small. Moreover, in the second level interaction,  while most function words and punctuation point to the utterance itself (i.e., $u$),  the word ``student'' indicates that information from ``i use all of my time for my education.'' is useful to recognize the relationship between the response candidate and the context. The example explains why DGMN works well from one perspective. 

\paragraph{Performance Analysis in Terms of Document Length.}
Finally, we study the relationship between the performance of DGMN and document length by binning text examples in both data into different buckets according to the document length. Table~\ref{tab:doc_lens} reports the evaluation results. On the PERSONA-CHAT data, both short profiles and long profiles lead to performance drop, while on the CMUDoG data, the longer the documents are, the better the performance of DGMN is. The reason behind the difference might be that profiles in the PERSONA-CHAT data are handcrafted by crowd workers, and thus semantics among different sentences are relatively independent, while documents in the CMUDoG data come from Wikipedia, and there is rich semantic overlap among sentences. Therefore, short profiles contain less useful information and long profiles contain more irrelevant information, and both will make the matching task more challenging. On the other hand, the longer a wiki document is, the more relevant information it can provide to the matching task.

\section{Related Work}
There are two groups of methods for building a chatbot. The first group learns response generation models under an encoder-decoder framework \cite{shangL2015neural,vinyals2015neural} with extensions to suppress generic responses \cite{li2015diversity,mou2016sequence,xing2017topic,tao2018get}. The second group learns a matching model of a human input and a response candidate for response selection. Along this line, early work assumes that the input is a single message~\cite{wang2013dataset,hu2014convolutional}. Recently, conversation history is taken into account in matching. Representative methods include the dual LSTM model~\cite{lowe2015ubuntu}, the deep learning to respond architecture~\cite{rui2018learning}, the multi-view matching model~\cite{zhou2016multi}, the sequential matching network~\cite{wu2017sequential}, the deep attention matching network~\cite{zhou2018multi}, and the multi-representation fusion network~\cite{tao2019multi}.  
Our work belongs to the second group. The major difference we make is that in addition to conversation contexts, we also incorporate external documents as a kind of background knowledge into matching.

Before us, a few recent studies have considered grounding open domain dialogues with external knowledge. For example, \newcite{ghazvininejad2018knowledge} generalize the vanilla Seq2seq model by conditioning responses on both conversation history and external ``facts''. \newcite{zhang2018personalizing} release a persona-based conversation data set where profiles created by crowd workers constrain speakers' personas in conversation. \newcite{mazare2018training} further increase the scale of the persona-chat data with conversations extracted from Reddit. \newcite{zhou2018dataset} publish a data set in which conversations are grounded in movie-related articles from Wikipedia. \newcite{dinan2018wizard} release another document-grounded data set with wiki articles covering broader topics. In this work, we study grounding retrieval-based open domain dialog systems with background documents and focus on building a powerful matching model with advanced neural architectures. On the persona-chat data published in \newcite{zhang2018personalizing} and the document-grounded conversation data set published in \newcite{zhou2018dataset}, the model improves upon state-of-the-art methods with large margins.    

\section{Conclusions}
We propose a document-grounded matching network to incorporate external knowledge into response selection for retrieval-based chatbots. Experimental results on two public data sets consistently show that the proposed model can significantly outperform state-of-the-art methods.

\section*{Acknowledgments}
This work was supported by the National Key Research and Development Program of China (No. 2017YFC0804001), the National Science Foundation of China (NSFC Nos. 61672058 and 61876196).

\bibliographystyle{named}
\bibliography{ijcai19}
\end{document}